\documentclass[]{fairmeta} %

\usepackage{amsmath,amsfonts,bm}

\def\eqref#1{equation~\ref{#1}}

\def\1{\bm{1}}

\DeclareMathAlphabet{\mathsfit}{\encodingdefault}{\sfdefault}{m}{sl}
\SetMathAlphabet{\mathsfit}{bold}{\encodingdefault}{\sfdefault}{bx}{n}

\usepackage[utf8]{inputenc}  %
\usepackage[T1]{fontenc}     %
\usepackage{hyperref}        %
\usepackage{url}             %
\usepackage{booktabs}        %
\usepackage{amsfonts}        %
\usepackage{nicefrac}        %
\usepackage{microtype}       %
\usepackage{graphicx}
\usepackage{multicol}
\usepackage{multirow}
\usepackage{xspace}
\usepackage{wrapfig}
\usepackage{enumitem}
\usepackage[dvipsnames,table]{xcolor}
\usepackage{colortbl}
\usepackage{caption}
\usepackage{lmodern} 
\usepackage{algorithm}
\usepackage{algorithmic}
\usepackage{subcaption}
\usepackage{wrapfig}
\usepackage{xspace}
\usepackage{xcolor} 
\usepackage{amsmath} 
\usepackage{multicol}
\usepackage{multirow}
\usepackage{pifont}
\usepackage{tabularx}
\newcolumntype{R}{>{\raggedleft\arraybackslash}X}
\definecolor{lightblue}{RGB}{220,235,250}
\definecolor{lightgray}{gray}{0.91}

\usepackage[most,skins,theorems]{tcolorbox}
\usepackage{listings}
\newtcolorbox{promptbox}[1][]{
  enhanced, breakable,
  colback=gray!1,      
  colframe=gray!60,    
  coltitle=black,      
  boxrule=2pt,
  arc=10pt,
  left=6pt, right=6pt, top=6pt, bottom=6pt,
  title={#1}, fonttitle=\bfseries,
  attach boxed title to top left={yshift*=-3mm},
  boxed title style={colback=gray!10}
}

\tcbset{
  aibox/.style={
    width=\linewidth,
    top=8pt,
    bottom=4pt,
    colback=inftythink-red!15,
    colframe=inftythink-red,
    colbacktitle=inftythink-red!90!black,
    enhanced,
    center,
    attach boxed title to top left={yshift=-0.1in,xshift=0.15in},
    boxed title style={boxrule=0pt,colframe=white,},
  }
}
\newtcolorbox{AIbox}[2][]{aibox,title=#2,#1}

\usepackage{amsthm}
\theoremstyle{plain}

\theoremstyle{definition}

\theoremstyle{remark}

\newcommand{\methodname}{UI-Zoomer\xspace}

\title{
UI-Zoomer: Uncertainty-Driven Adaptive Zoom-In for GUI Grounding
}

\author[1,*]{Fei Tang}
\author[1,*]{Bofan Chen}
\author[1]{Zhengxi Lu}
\author[1]{Tongbo Chen}
\author[2]{Songqin Nong}
\author[2]{Tao Jiang}
\author[2]{Wenhao Xu}
\author[1]{Weiming Lu}
\author[1]{Jun Xiao}
\author[1]{Yueting Zhuang}
\author[1 \dagger]{Yongliang Shen}

\affiliation[1]{Zhejiang University}
\affiliation[2]{Ant Group}

\contribution[*]{Equal Contribution}
\contribution[\dagger]{Corresponding authors}

\abstract{
GUI grounding, which localizes interface elements from screenshots given natural language queries, remains challenging for small icons and dense layouts.  
Test-time zoom-in methods improve localization by cropping and re-running inference at higher resolution, but apply cropping uniformly across all instances with fixed crop sizes, ignoring whether the model is actually uncertain on each case.
We propose \textbf{UI-Zoomer}, a training-free adaptive zoom-in framework that treats both the trigger and scale of zoom-in as a prediction uncertainty quantification problem.
A confidence-aware gate fuses spatial consensus among stochastic candidates with token-level generation confidence to selectively trigger zoom-in only when localization is uncertain. When triggered, an uncertainty-driven crop sizing module decomposes prediction variance into inter-sample positional spread and intra-sample box extent, deriving a per-instance crop radius via the law of total variance. 
Extensive experiments on ScreenSpot-Pro, UI-Vision, and ScreenSpot-v2 demonstrate consistent improvements over strong baselines across multiple model architectures, achieving gains of up to +13.4\%, +10.3\%, and +4.2\% respectively, with no additional training required.
}

\date{\today}
\metadata[Project Page]{\url{https://zju-real.github.io/UI-Zoomer}}
\metadata[Code]{\url{https://github.com/ZJU-REAL/UI-Zoomer}}
\correspondence{\email{syl@zju.edu.cn}}

\begin{document}

\maketitle

\section{Introduction}
\label{sec:intro}

Grounding natural language instructions to interface elements is a fundamental capability for autonomous GUI agents~\cite{uground,guig2,tang2025surveymllmbasedguiagents,xu2025aguvisunifiedpurevision,hong2024cogagentvisuallanguagemodel,lin2024showuivisionlanguageactionmodelgui,jiang2025appagentxevolvingguiagents,yang2023setofmarkpromptingunleashesextraordinary}. Despite significant progress through supervised fine-tuning and reinforcement learning~\cite{uitars15,xu2024aguvis,jedi,segui,uivenus},
models still fail systematically on small icons and dense layouts in complex interfaces~\cite{scrrenspotpro}. 

A natural remedy is \emph{test-time zoom-in scaling}: crop a region of the screenshot and re-run the model at higher effective resolution~\cite{dimo,regionfocus,nguyen2024improved,reguide}.While this paradigm has shown clear promise for fine-grained GUI localization~\cite{dimo,regionfocus}, a more fundamental question remains unaddressed: \textbf{which instances actually need zoom-in, and how much should we zoom?}

Existing zoom-in methods share two fundamental limitations. First, they apply cropping indiscriminately: \cite{dimo} zooms in unconditionally on every sample with a fixed scaling factor, while \cite{regionfocus} triggers zoom-in only upon execution errors, with no regard to whether the model is actually uncertain on the instance at hand. 
We show empirically that unconditional zoom-in on ScreenSpot-v2 degrades accuracy below the direct prediction baseline while significantly increasing latency (Table~\ref{tab:dimo_motivation}), as easy cases lose the global context the model was already exploiting. 
Second, all existing methods fix the crop window to a predetermined ratio~\cite{dimo,regionfocus,reguide}, regardless of whether candidates are tightly clustered or widely scattered, leaving the crop either too broad to improve resolution or too narrow to retain critical context.

\begin{figure}[t]
    \centering
    \includegraphics[width=\textwidth]{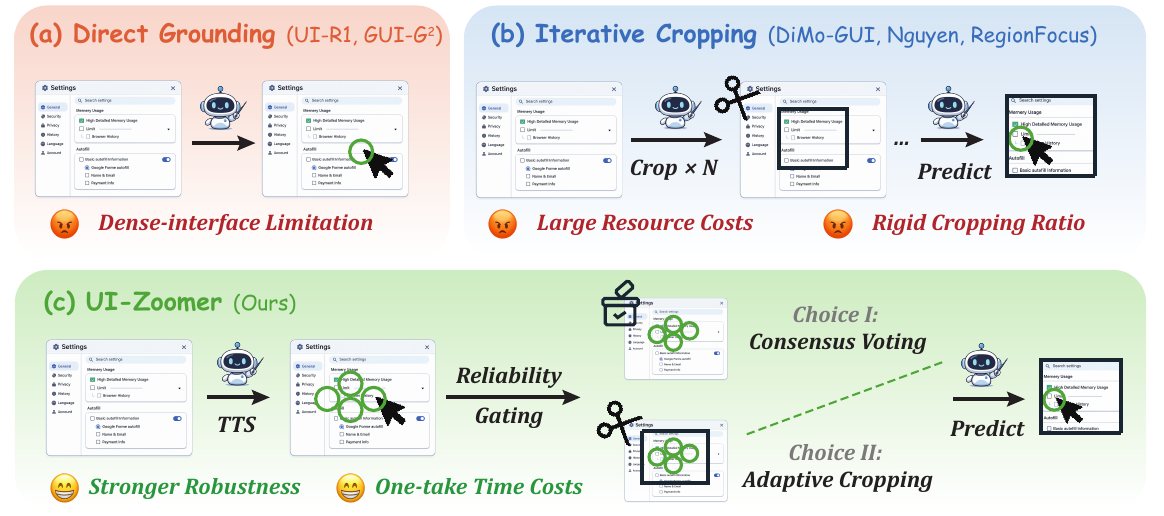}
    \caption{Comparison of GUI grounding paradigms. (a) Direct grounding methods struggle with dense interfaces. (b) Iterative cropping methods incur large resource costs and use rigid cropping ratios. (c) Our \methodname applies Test-Time Scaling (TTS) with reliability gating, adaptively choosing between consensus voting and adaptive cropping, achieving stronger robustness with one-take time costs.}
    \label{fig:motivation}
    \vspace{-4mm}
\end{figure}

\begin{wraptable}{r}{0.42\linewidth}

\centering
\small
\begin{tabular}{lcc}
\toprule
\textbf{Method} & \textbf{Avg Acc} & \textbf{Time} \\
\midrule
w/o DiMo-GUI & 81.84\% & 35:47 \\
w/ DiMo-GUI  & 77.20\% & 6:43:07 \\
\bottomrule
\end{tabular}
\caption{Accuracy and inference time of w/ and w/o iterative cropping on ScreenSpot-V2.}\label{tab:dimo_comparison}
\label{tab:dimo_motivation}

\end{wraptable}

The root cause is that these methods treat all instances uniformly, without consulting the model's own prediction behavior. 
Recent work shows that spatial agreement across stochastic samples correlates with localization reliability~\cite{guirc}, and that coordinate likelihoods near a predicted point follow a smooth Gaussian distribution in pixel space~\cite{reguide}, confirming that VLMs implicitly encode continuous spatial uncertainty. The variance of sampled predictions~\cite{safeground,guirc} thus encodes both whether the model is confused and over what spatial extent, which is precisely the information needed to gate zoom-in and size the crop window. This motivates a simple but previously unexplored principle: \textbf{zoom only when uncertain, and zoom by how much the predictions disagree}.

Building on this insights, we propose \textbf{\methodname}, a training-free adaptive zoom-in framework for GUI grounding. \methodname first draws $N$ stochastic candidates from the model and computes a reliability score by fusing spatial consensus with token-level confidence; instances that pass the gate are resolved immediately by consensus voting. For uncertain instances, the crop window is derived from the variance of candidate predictions decomposed into inter-sample positional spread and intra-sample box extent, yielding a per-instance radius that contracts for easy cases and expands for hard ones. A single deterministic re-inference pass on the resulting crop completes the refinement.

Extensive experiments on three widely-adopted GUI grounding benchmarks, ScreenSpot-Pro, UI-Vision, and ScreenSpot-v2, demonstrate that \methodname consistently improves over strong baselines, achieving gains of up to +13.4\%, +10.3\%, and +4.2\% respectively. Icon targets benefit more than text targets on average, consistent with the intuition that compact and semantically ambiguous elements profit most from high-resolution refinement. Ablations confirm the independent contribution of each component and the advantage of adaptive crop sizing over any fixed-ratio alternative.

Our contributions are threefold:
\begin{itemize}
\item We propose \methodname, a training-free adaptive zoom-in framework that frames the trigger and scale of zoom-in as a prediction uncertainty quantification problem.
\item \methodname comprises a confidence-aware gate that avoids unnecessary computation by routing only uncertain instances to refinement, and a Gaussian-based adaptive crop sizing module that derives the crop window from the variance of candidate predictions.
\item Extensive experiments on ScreenSpot-Pro, UI-Vision, and ScreenSpot-v2 demonstrate consistent improvements across four model architectures, with gains of up to +13.4\% on ScreenSpot-Pro.
\end{itemize}

\section{Related Work}

\subsection{GUI Grounding}

GUI grounding requires predicting the pixel coordinates of an interface element given a screenshot and a natural language instruction. Early work builds pipeline-based systems that chain OCR, icon detectors, and LLMs for planning and element selection~\cite{zhang2025ufo2desktopagentos,wang2024mobileagentv2mobiledeviceoperation,li2024appagentv2advancedagent,agashe2024agent,zhang2025ufo2desktopagentos,liu2024autoglmautonomousfoundationagents,yang2023setofmarkpromptingunleashesextraordinary,bai2021uibertlearninggenericmultimodal,xu2024aguvis,osatlas,tang2025thinktwiceclickonce,guiactor,agashe2025agent}. A second generation trains specialist VLMs end-to-end on large-scale GUI corpora, with models such as UGround~\cite{uground}, OS-Atlas~\cite{osatlas}, and UI-TARS~\cite{uitars15} demonstrating strong cross-platform generalization. More recently, reinforcement fine-tuning has emerged as a data-efficient alternative: methods including UI-R1~\cite{uir1}, GUI-G$^2$~\cite{guig2}, SE-GUI~\cite{segui}, and UI-Venus~\cite{uivenus} apply GRPO-style objectives with coordinate accuracy rewards, matching or exceeding SFT models trained on orders of magnitude more data. Despite these advances, all training-time approaches share a hard ceiling at high resolution: once a target element is too small to resolve in a standard forward pass, additional training provides diminishing returns~\cite{dimo,regionfocus}.

\subsection{Test-Time Scaling for GUI Grounding}

Test-time scaling improves model performance at inference without modifying parameters~\cite{snell2024scalingllmtesttimecompute}. In GUI grounding, the dominant paradigm is zoom-in inference: DiMo-GUI~\cite{dimo} applies iterative zoom-in with a fixed crop ratio; RegionFocus~\cite{regionfocus} triggers zoom-in upon execution errors; ReGUIDE~\cite{reguide} uses KDE over multiple predictions to identify a high-density crop center; Nguyen~\cite{nguyen2024improved} proposes successive iterative narrowing. A parallel thread exploits prediction consistency as a reliability signal: GUI-RC~\cite{guirc} constructs spatial voting grids over stochastic samples to identify consensus regions; SafeGround~\cite{safeground} derives calibrated uncertainty estimates from spatial dispersion with statistical guarantees; GUI-Eyes~\cite{guieyes} trains models via RL to actively decide when to invoke zoom tools. These methods either apply cropping regardless of per-instance confidence, or use consistency signals purely for voting without connecting them to crop sizing. UI-Zoomer unifies both perspectives by using prediction variance to simultaneously gate zoom-in and derive per-instance crop windows.

\section{Method}
\label{sec:method}

\subsection{Problem Setup}

Given a GUI screenshot $I \in \mathbb{R}^{H \times W \times 3}$ and a natural-language instruction $q$, we predict a click location $\hat{\mathbf{p}} \in [0,1]^2$ in normalized image coordinates. We represent each localization hypothesis as an axis-aligned bounding box $\mathbf{b} = [x_1, y_1, x_2, y_2]$ and define the click as its center:
\begin{equation}
  \hat{\mathbf{p}} = \left[\frac{x_1+x_2}{2},\ \frac{y_1+y_2}{2}\right].
  \label{eq:center}
\end{equation}

\begin{figure}[!t]
    \centering
    \includegraphics[width=0.95\textwidth]{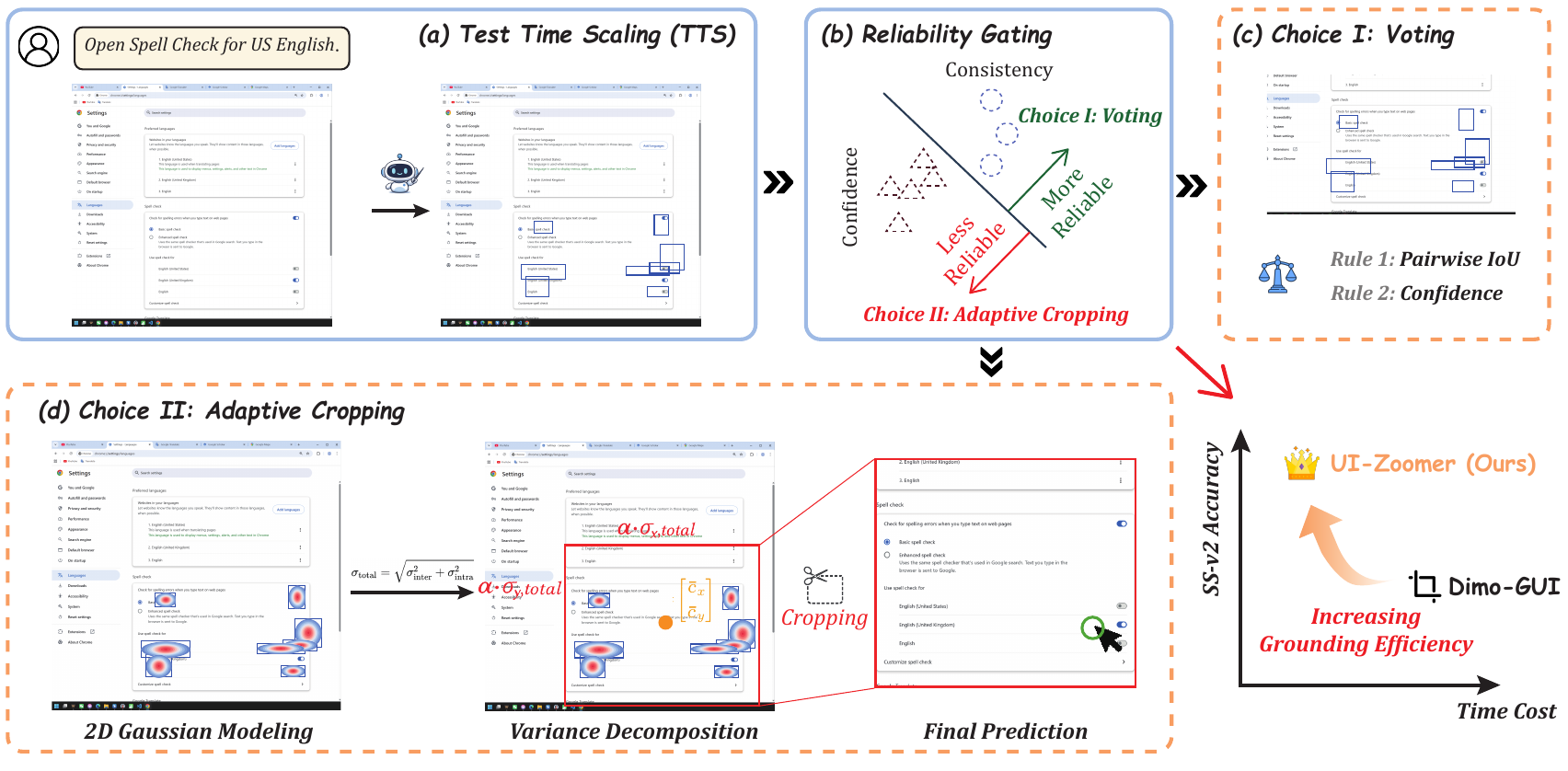}
    \caption{Overview of \methodname. (a) The model samples $N$ candidate predictions via Test-Time Scaling (TTS). (b) A reliability gate routes confident instances to consensus voting (Choice I) and uncertain ones to adaptive cropping (Choice II). (d) The crop window is derived from 2D Gaussian variance decomposition, enabling per-instance adaptive zoom-in.}
    \label{fig:method}

\end{figure}

As shown in Figure~\ref{fig:method}, UI-Zoomer proceeds in three stages: (1) global multi-sampling, (2) reliability gating, and (3) adaptive crop and zoom. The full procedure is summarized in Algorithm~\ref{alg:ui-zoomer}.

\begin{algorithm}[t]
\caption{UI-Zoomer}
\label{alg:ui-zoomer}
\begin{algorithmic}[1]
\REQUIRE Image $I$, instruction $q$, model $\mathcal{M}$, $N$, threshold $\tau$, scale $\gamma$, min crop $m$
\ENSURE Click point $\hat{\mathbf{p}} \in [0,1]^2$
\STATE $\{\mathbf{b}_i, c_i\}_{i=1}^N \leftarrow \mathrm{Sample}(\mathcal{M}, I, q; \ T{=}0.9)$
\hfill\textcolor{teal}{Stage 1: global multi-sampling}
\STATE Compute $C_{\mathrm{spatial}}$ (Eq.~3), $\bar{c}$ (Eq.~2), $S = C_{\mathrm{spatial}} + \bar{c}$
\hfill\textcolor{teal}{Stage 2: reliability gating}
\IF{$S > \tau$}
    \STATE \textbf{return} $\mathrm{center}(\mathbf{b}_{i^\star})$
    \hfill\textcolor{teal}{pass: consensus vote (Eq.~5)}
\ELSE
    \STATE $\boxed{\mu}, \boxed{\sigma} \leftarrow \mathrm{FilterAndDecompose}(\{\mathbf{b}_i\})$
    \hfill\textcolor{teal}{Stage 3: filter + variance decomp.}
    \STATE $(x_1^c, y_1^c, x_2^c, y_2^c) \leftarrow \mathrm{AdaptiveCrop}(\boxed{\mu}, \boxed{\sigma}; \ \gamma, m)$
    \hfill\textcolor{teal}{adaptive crop window (Eq.~10)}
    \STATE $\hat{\mathbf{b}} \leftarrow \mathcal{M}(\mathrm{Crop}(I, x_1^c, y_1^c, x_2^c, y_2^c); \ T{=}0)$
    \hfill\textcolor{teal}{zoom: deterministic re-inference}
    \STATE \textbf{return} $\mathrm{center}(\mathrm{MapBack}(\hat{\mathbf{b}}))$
    \hfill\textcolor{teal}{map back to global coords (Eq.~11)}
\ENDIF
\end{algorithmic}
\end{algorithm}

\subsection{Stage 1: Global Multi-Sampling}

We sample $N{=}8$ candidate boxes from $\mathcal{M}$ at temperature $T{=}0.9$ and discard invalid parses. For each valid candidate $i$ we record the predicted box $\mathbf{b}_i$ and estimate a scalar confidence from the geometric mean of token probabilities:
\begin{equation}
  c_i = \exp\!\left(\frac{1}{L_i}\sum_{t=1}^{L_i} \log p_{i,t}\right),
  \label{eq:conf}
\end{equation}
where $L_i$ is the sequence length and $p_{i,t}$ is the probability of the $t$-th token.

\subsection{Stage 2: Reliability Gating}
\label{sec:gate}

When candidates are consistent and confident, zoom-in is unnecessary and costly. We quantify this reliability through two complementary signals and use their combination to selectively trigger refinement.

\subsubsection{Spatial consensus.} We quantify cross-sample agreement by the mean pairwise IoU:
\begin{equation}
  C_\text{spatial} = \frac{1}{N(N-1)} \sum_{i \neq j} \mathrm{IoU}(\mathbf{b}_i, \mathbf{b}_j).
  \label{eq:cspatial}
\end{equation}

\subsubsection{Gating score.} We combine spatial consensus with average token confidence:
\begin{equation}
  S = C_\text{spatial} + \bar{c}, \qquad \bar{c} = \frac{1}{N}\sum_{i=1}^{N} c_i.
  \label{eq:gating}
\end{equation}
The two signals are complementary: $C_\text{spatial}$ is sensitive to positional scatter while $\bar{c}$ reflects sharpness of the predictive distribution over coordinate tokens. When $S > \tau$, we trust the global predictions and return immediately.

\subsubsection{Consensus voting.} We select the candidate with the most peer support, breaking ties by confidence:
\begin{equation}
  v_i = \sum_{j \neq i} \mathbb{I}[\mathrm{IoU}(\mathbf{b}_i, \mathbf{b}_j) > 0.5], \qquad
  i^\star = \arg\max_i\,(v_i,\ c_i).
  \label{eq:vote}
\end{equation}

\subsection{Stage 3: Uncertainty-Driven Adaptive Crop}
\label{sec:crop}

When $S \leq \tau$, candidates are unreliable and zoom-in is warranted. Rather than using a fixed crop ratio, we derive the crop window directly from the variance of the candidate set.

\subsubsection{Outlier filtering.} A small number of erratic samples can inflate the estimated variance and produce an oversized crop. We therefore discard outliers by retaining only the $K = \lfloor 0.75N \rfloor$ candidates whose centers lie closest to the median center $\tilde{\mathbf{z}}$:
\begin{equation}
  d_i = \|\mathbf{z}_i - \tilde{\mathbf{z}}\|_2, \qquad
  \mathcal{K} = \mathop{\arg\mathrm{topK}}_{i}\ \{-d_i\},
  \label{eq:filter}
\end{equation}
where $\mathbf{z}_i$ denotes the center of $\mathbf{b}_i$. We compute subsequent statistics over $\mathcal{K}$.

\subsubsection{Variance decomposition.} We model the unknown target location $\mathbf{Z}$ as a latent random variable and apply the law of total variance coordinate-wise:
\begin{equation}
  \mathrm{Var}(\mathbf{Z}) =
  \underbrace{\mathrm{Var}\!\left(\mathbb{E}[\mathbf{Z} \mid I]\right)}_{\mathbf{v}_\text{inter}}
  +
  \underbrace{\mathbb{E}\!\left[\mathrm{Var}(\mathbf{Z} \mid I)\right]}_{\mathbf{v}_\text{intra}}.
  \label{eq:var_decomp}
\end{equation}
The inter-sample term captures positional disagreement across draws:
\begin{equation}
  \mathbf{v}_\text{inter} = \frac{1}{K}\sum_{i \in \mathcal{K}} (\mathbf{z}_i - \boldsymbol{\mu})^{\odot 2},
  \qquad \boldsymbol{\mu} = \frac{1}{K}\sum_{i \in \mathcal{K}} \mathbf{z}_i.
  \label{eq:inter}
\end{equation}
The intra-sample term encodes the predicted scale of each element. Treating each box as a Gaussian spanning ${\pm}2\sigma$ of its width and height:
\begin{equation}
  \mathbf{v}_\text{intra} = \frac{1}{K}\sum_{i \in \mathcal{K}} \left(\frac{\mathbf{s}_i}{4}\right)^{\!\odot 2},
  \label{eq:intra}
\end{equation}
where $\mathbf{s}_i = [s_{ix}, s_{iy}]^\top$ is the width and height of $\mathbf{b}_i$. The two terms are complementary: $\mathbf{v}_\text{inter}$ expands the crop when candidates disagree on position; $\mathbf{v}_\text{intra}$ ensures the crop is at least as large as the predicted element even when candidates coincide.

\subsubsection{Crop window.} We set the crop radius as $\mathbf{r} = \gamma\boldsymbol{\sigma}$, where $\boldsymbol{\sigma} = \sqrt{\mathbf{v}_\text{inter} + \mathbf{v}_\text{intra}}$. To avoid degenerate crops and aspect-ratio distortions, we impose a minimum side length $m$ and squarify:

\begin{equation}
s = \max(2r_x,2r_y,m),
\quad 
[x_1^c,y_1^c,x_2^c,y_2^c]
=
[\mu_x-\tfrac{s}{2},\;\mu_y-\tfrac{s}{2},\;\mu_x+\tfrac{s}{2},\;\mu_y+\tfrac{s}{2}].
\label{eq:crop}
\end{equation}
If the window extends beyond image boundaries, we shift it inward while preserving its size.

\subsubsection{Zoom and map back.} We crop $I$ to this window, resize it to the model's resolution budget, and run a single deterministic pass ($T{=}0$) to obtain a refined box $\hat{\mathbf{b}}$ in crop coordinates. We map it back to global normalized coordinates via:
\begin{equation}
  x = \frac{x_1^c + \hat{x}\, w_c}{W}, \qquad y = \frac{y_1^c + \hat{y}\, h_c}{H},
  \label{eq:mapback}
\end{equation}
where $w_c = x_2^c - x_1^c$ and $h_c = y_2^c - y_1^c$. If refinement produces an invalid box, we fall back to the most confident global candidate.

\section{Experiments}
\label{sec:exp}

\subsection{Setup}

\subsubsection{Benchmarks.}

We evaluate on three benchmarks spanning different difficulty regimes.
\textbf{ScreenSpot-Pro}~\cite{scrrenspotpro} targets 4K professional desktop environments across 23 applications, with unusually small and dense targets.
\textbf{ScreenSpot-v2}~\cite{osatlas} is a multi-platform benchmark covering mobile, desktop, and web interfaces with 1,200+ instructions.
\textbf{UI-Vision}~\cite{uivision} covers fine-grained desktop grounding across 83 real-world applications, including element grounding, layout grounding, and action prediction.
Following prior work~\cite{screenspot,scrrenspotpro}, we report click accuracy: a prediction is correct if the output point falls within the ground-truth bounding box.

\subsubsection{Models.}
We evaluate our method using two categories of base models: (1) \textbf{general-purpose VLMs}, i.e., \textbf{Qwen2.5-VL-7B}~\cite{qwen25vl}, an open-source multimodal model pretrained on large-scale data; and (2) \textbf{GUI-specific VLMs}, including \textbf{UI-Venus-7B}, \textbf{UI-Venus-72B}~\cite{uivenus}, and \textbf{GUI-G$^2$-7B}~\cite{guig2}, which are tailored for GUI understanding and grounding. 
Notably, both UI-Venus and GUI-G$^2$ are further enhanced with reinforcement learning, leading to stronger task-specific alignment for UI interaction and more reliable GUI grounding behaviors. 

\subsubsection{Implementation.} 
All evaluations are conducted on 4 NVIDIA RTX 4090D 24G GPUs.
We use the vLLM engine with a context length of 16{,}384 tokens. We sample $N{=}8$ candidates at temperature $T{=}0.9$ and set the minimum crop side to $m{=}512$ pixels.
The gating threshold $\tau$ and Gaussian scale $\gamma$ are tuned per model-benchmark pair; for UI-Venus-7B on ScreenSpot-Pro we use $\tau{=}1.0$ and $\gamma{=}2.5$.

\subsection{Main Results}

\begin{table}[!t]
\centering
\resizebox{\textwidth}{!}{%
\begin{tabular}{c|c|ccc|ccc|ccc|ccc}
\toprule
\multirow{2}{*}{\textbf{Benchmark}} &
\multirow{2}{*}{\textbf{Methods}} &
\multicolumn{3}{c|}{\textbf{Category 1}} & 
\multicolumn{3}{c|}{\textbf{Category 2}} & 
\multicolumn{3}{c|}{\textbf{Category 3}} & 
\multicolumn{3}{c}{\textbf{Overall}} \\
 & & text & icon & avg & text & icon & avg & text & icon & avg & text & icon & avg \\
\midrule
\multirow{3}{*}{\textbf{ScreenSpot-v2}}
 & UI-Venus-7B & 
99.0 & 90.1 & 95.2 & 
97.9 & 89.3 & 94.3 & 
94.9 & 89.7 & 92.5 & 
97.4 & 89.7 & 94.0 \\
 & \textbf{+ \methodname} & 
\textbf{98.6} & \textbf{90.5} & \textbf{95.2} & 
\textbf{99.0} & \textbf{92.9} & \textbf{96.4} & 
\textbf{95.7} & \textbf{90.6} & \textbf{93.4} & 
\textbf{97.8} & \textbf{91.2} & \textbf{94.9} \\
 & \textcolor{red}{$\Delta$ \textit{Improvement}} & 
\textcolor{red}{-0.4} & \textcolor{red}{+0.4} & \textcolor{red}{+0.0} & 
\textcolor{red}{+1.1} & \textcolor{red}{+3.6} & \textcolor{red}{+2.1} & 
\textcolor{red}{+0.8} & \textcolor{red}{+0.9} & \textcolor{red}{+0.9} & 
\textcolor{red}{+0.4} & \textcolor{red}{+1.5} & \textcolor{red}{+0.9} \\
\midrule
\multirow{3}{*}{\textbf{UI-Vision}}
 & UI-Venus-7B & 
68.5 & 23.8 & 33.3 & 
65.5 & 19.9 & 29.7 & 
38.7 & 7.8 & 11.4 & 
60.6 & 16.5 & 24.4 \\
 & \textbf{+ \methodname} & 
\textbf{80.5} & \textbf{32.3} & \textbf{42.5} & 
\textbf{74.7} & \textbf{30.6} & \textbf{40.1} & 
\textbf{55.9} & \textbf{15.1} & \textbf{19.7} & 
\textbf{72.7} & \textbf{25.2} & \textbf{33.7} \\
 & \textcolor{red}{$\Delta$ \textit{Improvement}} & 
\textcolor{red}{+12.0} & \textcolor{red}{+8.5} & \textcolor{red}{+9.2} & 
\textcolor{red}{+9.2} & \textcolor{red}{+10.7} & \textcolor{red}{+10.4} & 
\textcolor{red}{+17.2} & \textcolor{red}{+7.3} & \textcolor{red}{+8.3} & 
\textcolor{red}{+12.1} & \textcolor{red}{+8.7} & \textcolor{red}{+9.3} \\
\bottomrule
\end{tabular}%
}
\caption{Performance of UI-Venus-7B with and without \methodname on ScreenSpot-v2 (Mobile / Desktop / Web) and UI-Vision (Basic / Functional / Spatial).}
\label{tab:uiv_ssv2_merge_results}
\vspace{-2mm}
\end{table}

\begin{table}[!t]
\centering
\resizebox{\textwidth}{!}{%
\begin{tabular}{l|cc|cc|cc|cc|cc|cc|ccc}
\toprule
\multirow{2}{*}{\textbf{Methods}} &
\multicolumn{2}{c|}{\textbf{Development}} & 
\multicolumn{2}{c|}{\textbf{Creative}} & 
\multicolumn{2}{c|}{\textbf{CAD}} & 
\multicolumn{2}{c|}{\textbf{Scientific}} & 
\multicolumn{2}{c|}{\textbf{Office}} & 
\multicolumn{2}{c|}{\textbf{OS}} & 
\multicolumn{3}{c}{\textbf{Overall}} \\

& text & icon & text & icon & text & icon & text & icon & text & icon & text & icon & text & icon & avg \\
\midrule
\rowcolor{gray!15}
\multicolumn{16}{l}{\textit{Proprietary Methods}} \\
GPT-4o~\cite{gpt4o} & 1.3 & 0.0 & 1.0 & 0.0 & 2.0 & 0.0 & 2.1 & 0.0 & 1.1 & 0.0 & 0.0 & 0.0 & 1.3 & 0.0 & 0.8 \\
Claude-3.7-Sonnet~\cite{claude37sonnet} & - & - & - & - & - & - & - & - & - & - & - & - & - & - & 27.7 \\
Seed-1.5-VL~\cite{seed15vl} & - & - & - & - & - & - & - & - & - & - & - & - & - & - & 60.9 \\

\midrule
\rowcolor{gray!15}
\multicolumn{16}{l}{\textit{General Open-source Models}} \\
OS-Atlas-7B~\cite{osatlas} & 33.1 & 1.4 & 28.8 & 2.8 & 12.2 & 4.7 & 37.5 & 7.3 & 33.9 & 5.7 & 27.1 & 4.5 & 28.1 & 4.0 & 18.9 \\
Qwen2.5-VL-3B~\cite{qwen25vl} & 38.3 & 3.4 & 40.9 & 4.9 & 22.3 & 6.3 & 44.4 & 10.0 & 48.0 & 17.0 & 33.6 & 4.5 & 37.8 & 6.6 & 25.9 \\
UGround-7B~\cite{uground} & - & - & - & - & - & - & - & - & - & - & - & - & - & - & 31.1 \\
UGround-72B & - & - & - & - & - & - & - & - & - & - & - & - & - & - & 34.5 \\
UI-TARS-7B & 58.4 & 12.4 & 50.0 & 9.1 & 20.8 & 9.4 & 63.9 & 31.8 & 63.3 & 20.8 & 30.8 & 16.9 & 47.8 & 16.2 & 35.7 \\
UI-TARS-72B & 63.0 & 17.3 & 57.1 & 15.4 & 18.8 & 12.5 & 64.6 & 20.9 & 63.3 & 26.4 & 42.1 & 15.7 & 50.9 & 17.5 & 38.1 \\
Jedi-7B~\cite{jedi} & 42.9 & 11.0 & 50.0 & 11.9 & 38.0 & 14.1 & 72.9 & 25.5 & 75.1 & 47.2 & 33.6 & 16.9 & 52.6 & 18.2 & 39.5 \\
Qwen2.5-VL-32B & 74.0 & 21.4 & 61.1 & 13.3 & 38.1 & 15.6 & 78.5 & 29.1 & 76.3 & 37.7 & 55.1 & 27.0 & 63.2 & 22.5 & 47.6 \\

\midrule
\rowcolor{gray!15}
\multicolumn{16}{l}{\textit{Reinforcement Learning Methods}} \\
UI-TARS-1.5~\cite{uitars15} & - & - & - & - & - & - & - & - & - & - & - & - & - & - & 61.6 \\
GTA1-7B~\cite{gta1} & 53.3 & 17.2 & 66.9 & 20.7 & 62.6 & 18.2 & 76.4 & 31.8 & 82.5 & 50.9 & 48.6 & 25.9 & 65.5 & 25.2 & 50.1 \\
UI-R1-E-3B~\cite{uir1} & 46.1 & 6.9 & 41.9 & 4.2 & 37.1 & 12.5 & 56.9 & 21.8 & 65.0 & 26.4 & 32.7 & 10.1 & - & - & 33.5  \\
UI-S1-7B~\cite{u1s1} & - & - & - & - & - & - & - & - & - & - & - & - & - & - & 30.6 \\
SE-GUI-7B~\cite{segui} & 51.3 & 42.2 & 68.2 & 19.3 & 57.6 & 9.1 & 75.0 & 28.2 & 78.5& 43.4 & 49.5 & 25.8 & 63.5 & 21.0 & 47.3 \\

\midrule
\rowcolor{gray!15}
\multicolumn{16}{l}{\textit{Test Scaling Methods}} \\
DiMo-GUI~\cite{dimo} & 66.9 & 21.4 & 60.6 & 21.7 & 50.3 & 14.1 & 68.1 & 21.8 & 80.8 & 52.8 & 69.2 & 28.1 & 65.2 & 24.5 & 49.7 \\
RegionFocus~\cite{regionfocus} & 53.2 & 3.4 & 42.9 & 4.9 & 28.4 & 3.1 & 56.9 & 10.9 & 59.9 & 24.5 & 41.1 & 15.7 & 46.6 & 8.8 & 32.1 \\
GUI-RC~\cite{guirc} & - & - & - & - & - & - & - & - & - & - & - & - & - & - & 24.0  \\ 
UI-Venus-7B [pass@4] & 77.9 & 29.0 & 68.0 & 19.6 & 66.0 & 25.00 & 79.2 & 26.4 & 83.2 & 37.7 & 58.9 & 25.8 & 72.6 & 26.2 & 54.8  \\
UI-Venus-7B [pass@8] & 81.2 & 32.4 & 70.1 & 24.5 & 69.5 & 28.1 & 81.3 & 29.1 & 87.0 & 43.4 & 66.4 & 27.0 & 75.8 & 29.6 & 58.2  \\
\midrule
\rowcolor{gray!15}
\multicolumn{16}{l}{\textit{Our method}} \\
Qwen2.5-VL-7B & 
48.7 & 2.1 & 
32.0 & 4.9 & 
24.4 & 4.7 & 
51.4 & 7.3 & 
53.7 & 18.9 & 
38.3 & 10.1 & 
40.6 & 6.6  & 27.6\\

\textbf{+ \methodname} & 
\textbf{63.6} & \textbf{17.9} & 
\textbf{45.7} & \textbf{14.0} & 
\textbf{51.3} & \textbf{14.1} & 
\textbf{47.2} & \textbf{20.0} & 
\textbf{66.3} & \textbf{34.0} & 
\textbf{49.5} & \textbf{28.1} & 
\textbf{54.0} & \textbf{19.9} & \textbf{41.0}\\

\textcolor{red}{$\Delta$ \textit{Improvement}} & 
\textcolor{red}{+14.9} & \textcolor{red}{+15.8} & 
\textcolor{red}{+13.7} & \textcolor{red}{+9.1} & 
\textcolor{red}{+26.9} & \textcolor{red}{+9.4} & 
\textcolor{red}{-4.2} & \textcolor{red}{+12.7} & 
\textcolor{red}{+12.6} & \textcolor{red}{+15.1} & 
\textcolor{red}{+11.2} & \textcolor{red}{+18.0} & 
\textcolor{red}{+13.4} & \textcolor{red}{+13.3} & \textcolor{red}{+13.4} \\

\midrule

GUI-G$^2$-7B~\cite{guig2} & 
67.5 & 24.1 & 
59.9 & 16.1 & 
55.3 & 20.3 & 
75.7 & 28.2 & 
75.8 & 39.6 & 
50.5 & 20.2 & 
64.4 & 23.3 & 48.7 \\

\textbf{+ \methodname} & 
\textbf{79.9} & \textbf{38.6} & 
\textbf{68.0} & \textbf{26.6} & 
\textbf{77.7} & \textbf{34.4} & 
\textbf{82.6} & \textbf{36.4} & 
\textbf{84.3} & \textbf{60.4} & 
\textbf{65.4} & \textbf{38.2} & 
\textbf{76.7} & \textbf{36.8} & \textbf{61.4} \\

\textcolor{red}{$\Delta$ \textit{Improvement}} & 
\textcolor{red}{+12.3} & \textcolor{red}{+14.5} & 
\textcolor{red}{+8.1} & \textcolor{red}{+10.5} & 
\textcolor{red}{+22.3} & \textcolor{red}{+14.1} & 
\textcolor{red}{+7.0} & \textcolor{red}{+8.2} & 
\textcolor{red}{+8.4} & \textcolor{red}{+20.8} & 
\textcolor{red}{+15.0} & \textcolor{red}{+18.0} & 
\textcolor{red}{+12.3} & \textcolor{red}{+13.4} & \textcolor{red}{+12.7} \\

\midrule

UI-Venus-7B~\cite{uivenus} & 
72.7 & 22.8 & 
62.4 & 15.4 & 
58.9 & 21.9 & 
74.3 & 26.4 & 
78.7 & 35.9 & 
50.5 & 23.6 & 
66.7 & 22.9 & 50.0 \\

\textbf{+ \methodname} & 
\textbf{80.5} & \textbf{37.2} & 
\textbf{70.1} & \textbf{31.5} & 
\textbf{77.2} & \textbf{34.4} & 
\textbf{82.6} & \textbf{30.0} & 
\textbf{88.8} & \textbf{50.9} & 
\textbf{67.3} & \textbf{37.1} & 
\textbf{78.1} & \textbf{35.4} & \textbf{61.8} \\

\textcolor{red}{$\Delta$ \textit{Improvement}} & 
\textcolor{red}{+7.8} & \textcolor{red}{+14.4} & 
\textcolor{red}{+7.7} & \textcolor{red}{+16.1} & 
\textcolor{red}{+18.3} & \textcolor{red}{+12.5} & 
\textcolor{red}{+8.3} & \textcolor{red}{+3.6} & 
\textcolor{red}{+10.1} & \textcolor{red}{+15.0} & 
\textcolor{red}{+16.8} & \textcolor{red}{+13.5} & 
\textcolor{red}{+11.4} & \textcolor{red}{+12.5} & \textcolor{red}{+11.8} \\

\midrule

UI-Venus-72B & 
80.5 & 32.4 & 
70.1 & 32.9 & 
63.5 & 29.7 & 
75.0 & 39.1 & 
83.7 & 49.1 & 
73.8 & 34.8 & 
74.0 & 35.3 & 59.2 \\

\textbf{+ \methodname} & 
\textbf{85.7} & \textbf{42.1} & 
\textbf{75.1} & \textbf{44.8} & 
\textbf{76.1} & \textbf{40.6} & 
\textbf{84.0} & \textbf{42.7} & 
\textbf{86.5} & \textbf{69.8} & 
\textbf{83.2} & \textbf{48.3} & 
\textbf{81.3} & \textbf{46.0} & \textbf{67.8} \\

\textcolor{red}{$\Delta$ \textit{Improvement}} & 
\textcolor{red}{+5.2} & \textcolor{red}{+9.7} & 
\textcolor{red}{+5.0} & \textcolor{red}{+11.9} & 
\textcolor{red}{+12.6} & \textcolor{red}{+10.9} & 
\textcolor{red}{+9.0} & \textcolor{red}{+3.6} & 
\textcolor{red}{+2.8} & \textcolor{red}{+20.7} & 
\textcolor{red}{+9.4} & \textcolor{red}{+13.5} & 
\textcolor{red}{+7.3} & \textcolor{red}{+10.7} & \textcolor{red}{+8.6} \\

\bottomrule
\end{tabular}%
}
\caption{Performance comparison on ScreenSpot-Pro across four models: 
Qwen2.5-VL-7B, GUI-G$^2$-7B, UI-Venus-7B, and UI-Venus-72B. 
For a fair comparison, RegionFocus is evaluated using Qwen2.5-VL-7B 
as the backbone.}
\label{tab:screenspot_pro}
\end{table}

Table~\ref{tab:screenspot_pro} reports results on ScreenSpot-Pro; UI-Vision and ScreenSpot-v2 results appear in Table~\ref{tab:uiv_ssv2_merge_results} (Full results are provided in \textbf{Appendix Tables}~\ref{tab:ui_vision_results} and~\ref{tab:ssv2_results}). \methodname consistently improves all four models across all three benchmarks, with average gains of up to \textbf{+13.4\%}, \textbf{+10.3\%}, and \textbf{+4.2\%} on ScreenSpot-Pro, UI-Vision, and ScreenSpot-v2 respectively.

\paragraph{Zoom-in is most effective where resolution matters most.}
Gains are largest on ScreenSpot-Pro, the highest-resolution benchmark, and smallest on ScreenSpot-v2, which covers standard-resolution mobile and web interfaces. Within ScreenSpot-Pro, icon targets benefit more than text targets across all models (+12.5\% vs.\ +11.1\%), consistent with the intuition that compact and semantically ambiguous elements are most limited by resolution in a single forward pass.

\paragraph{Adaptive zoom outperforms both naive sampling and prior test-time methods.}
Compared to naive sampling baselines (UI-Venus-7B pass@4: 54.84\%, pass@8: 58.19\%), UI-Zoomer reaches 61.8\% at a comparable inference budget. It also substantially outperforms the prior zoom-in method RegionFocus~\cite{regionfocus} (32.1\%), which applies cropping unconditionally with a fixed ratio. Against RL-trained methods, UI-Zoomer with UI-Venus-7B surpasses UI-S1-7B (30.6\%) by \textbf{+31.2\%} and GTA1-7B (50.1\%) by \textbf{+11.7\%} on ScreenSpot-Pro, showing that uncertainty-driven zoom-in provides gains complementary to train-time optimization.

\begin{figure}
    \centering
    \includegraphics[width=0.88\textwidth]{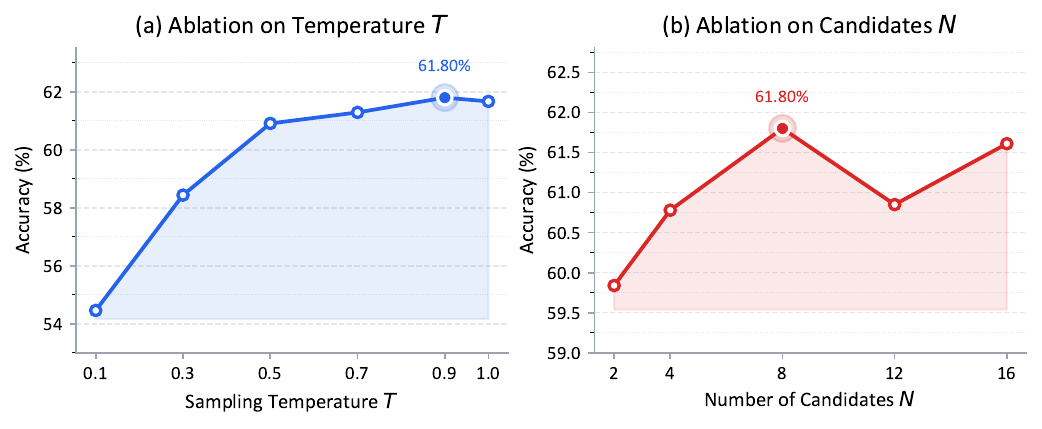}
    \caption{Ablation on sampling temperature $T$ (left) and number of candidates $N$ (right)
    on ScreenSpot-Pro.}
    \label{fig:TN_ablation}
\end{figure}

\section{Ablation Study}
\label{sec:ablation}

We conduct systematic ablation experiments on ScreenSpot-Pro with UI-Venus-7B
to validate the design of each component in \methodname.

\begin{table}[t]
\centering
\small
\begin{minipage}[t]{0.48\columnwidth}
\centering
\begin{tabular}{lc}
\toprule
\textbf{Variance Component} & \textbf{Accuracy (\%)} \\
\midrule
($\sigma_{intra}$) only & 60.97 \\
($\sigma_{inter}$) only & 61.42 \\
\textbf{($\sigma_{total} = \sigma_{intra} + \sigma_{inter}$)} & \textbf{61.80} \\
\bottomrule
\end{tabular}
\caption{Ablation study of variance components for adaptive zooming on ScreenSpot Pro. }
\label{tab:ablation_variance}
\end{minipage}
\hfill
\begin{minipage}[t]{0.48\columnwidth}
\centering
\begin{tabular}{lcc}
\toprule
\textbf{Components} & \textbf{$\tau$=0.6} & \textbf{$\tau$=1.0} \\
\midrule
$C_{spatial}$ only & 53.90\% & 60.81\% \\
$avg\_conf$ only & 50.22\% & 61.10\% \\
$C_{spatial} + avg\_conf$ & 57.56\% & \textbf{61.80\%} \\
\bottomrule
\end{tabular}
\caption{Ablation study of Gating Score components for uncertainty evaluation on ScreenSpot Pro. }
\label{tab:gating_conpone_ablation}
\end{minipage}
\vspace{-4mm}
\end{table}

\paragraph{Combining spatial consistency and average token confidence significantly improves the gating performance.} The gating score $S$ combines spatial consistency $C_{\text{spatial}}$ and average token confidence $avg\_conf$. As shown in Table~\ref{tab:gating_conpone_ablation}, using $C_{\text{spatial}}$ alone results in $60.81\%$ accuracy, while $avg\_conf$ alone achieves $61.10\%$. When both components are combined, the accuracy increases to 61.80\%, demonstrating the effectiveness of their combination. The complementarity of these two signals is evident from their distributional properties: $C_{\text{spatial}}$ shows a broad, spread distribution, whereas $avg\_conf$ is more concentrated (see Figure~\ref{fig:gating_hist}). This indicates that $C_{\text{spatial}}$ captures spatial variability, while $avg\_conf$ focuses on token-level certainty. By combining these signals, we can more effectively discriminate between uncertain samples, leading to a more reliable gating mechanism.

\paragraph{Decomposing crop uncertainty into intra-sample and inter-sample variance improves crop sizing.} \methodname decomposes crop uncertainty into intra-sample variance $\mathbf{v}_{\text{intra}}$ (box extent) and inter-sample variance $\mathbf{v}_{\text{inter}}$ (center disagreement across samples). As shown in Table~\ref{tab:ablation_variance}, both individual terms outperform the baseline, highlighting that each term captures a distinct and useful aspect of prediction uncertainty. $\mathbf{v}_{\text{intra}}$ encodes the predicted object scale, providing a lower bound on how large the crop should be, while $\mathbf{v}_{\text{inter}}$ reflects cross-sample positional spread, dynamically expanding the crop when candidates disagree. Since these two sources of uncertainty are complementary and not redundant, combining them results in a more complete characterization of ambiguity, leading to the best crop sizing, achieving 61.80\%.

\paragraph{Adaptive crop sizing provides a clear advantage over fixed-ratio alternatives.} A critical design consideration is whether adaptive crop sizing offers a real benefit over simpler fixed-ratio methods. Table~\ref{tab:ablation_cropping_strategy} compares fixed-ratio crops at three different scales with our Gaussian adaptive strategy. Fixed-ratio crops are sensitive to the chosen ratio: a large ratio ($0.8$) retains too much background information ($55.22\%$ accuracy), while a smaller ratio ($0.3$) risks cutting off important contextual cues, still trailing our method ($61.35\%$ vs. $\mathbf{61.80\%}$). In contrast, our Gaussian crop dynamically adjusts the crop window to the actual spread of candidate locations, achieving the best accuracy without requiring manual tuning of the crop scale.

\begin{table}[t]
\centering
\small

\begin{minipage}[t]{0.49\columnwidth}
\centering
\begin{tabular}{lc}
\toprule
\textbf{Boundary Strategy} & \textbf{Accuracy (\%)} \\
\midrule
Shrink & 58.47 \\
Clip & 60.25 \\
\textbf{Shift} & \textbf{61.80} \\
\bottomrule
\end{tabular}
\caption{Ablation study of crop box boundary handling strategies for UI-Venus-7B on ScreenSpot-Pro.}
\label{tab:ablation_boundary}
\end{minipage}
\hfill
\begin{minipage}[t]{0.49\columnwidth}
\centering
\begin{tabular}{lc}
\toprule
\textbf{Removal Ratio} & \textbf{Accuracy (\%)} \\
\midrule
50\% & 60.37 \\
\textbf{25\%} & \textbf{61.80} \\
0\%  & 60.03 \\
\bottomrule
\end{tabular}
\caption{Ablation study of outlier removal ratio for UI-Venus-7B on ScreenSpot-Pro.}
\label{tab:ablation_denoising}
\end{minipage}
\end{table}

\paragraph{Shifting the crop window inward yields the best performance.} When the computed crop window extends beyond the image boundaries, three strategies are possible: \textit{Shrink} (reduce window size), \textit{Clip} (hard-clip to the image edge), or \textit{Shift} (translate the window inward while preserving its size). As shown in Table~\ref{tab:ablation_boundary}, the \textit{Shift} strategy achieves 61.80\%, outperforming both \textit{Clip} (60.25\%) and \textit{Shrink} (58.47\%). Both \textit{Shrink} and \textit{Clip} alter the effective crop area, potentially losing important parts of the target region, whereas \textit{Shift} maintains the intended crop size, preserving the spatial context needed for accurate grounding.

\paragraph{Retaining the top 75\% of candidates yields the best accuracy.} Sampled candidates can contain spatial outliers that inflate the crop window, reducing effective resolution. To mitigate this, we retain only the top-$\rho$ fraction of candidates closest to the median center before fitting the Gaussian. As shown in Table~\ref{tab:ablation_denoising}, $\rho = 75\%$ achieves the best accuracy (61.80\%), balancing the removal of noisy predictions ($\rho = 50\%$: 60.37\%) with the retention of all candidates, including unfiltered outliers ($\rho = 100\%$: 60.03\%).
\begin{table}[t]
\centering
\small

\begin{minipage}[t]{0.47\columnwidth}
\centering
\begin{tabular}{lc}
\toprule
\textbf{Strategy} & \textbf{Accuracy (\%)} \\
\midrule
Fixed Ratio = 0.8 & 55.22 \\
Fixed Ratio = 0.5 & 59.58 \\
Fixed Ratio = 0.3 & 61.35 \\
\midrule
\textbf{\methodname} & \textbf{61.80} \\
\bottomrule
\end{tabular}
\caption{Ablation study of cropping strategy for UI-Venus-7B on ScreenSpot-Pro.}
\label{tab:ablation_cropping_strategy}
\end{minipage}
\hfill
\begin{minipage}[t]{0.47\columnwidth}
\centering
\begin{tabular}{lc}
\toprule
\textbf{Strategy} & \textbf{Accuracy (\%)} \\
\midrule
w/o Square Crop & 60.56 \\
w/ Square Crop  & \textbf{61.80} \\
\midrule
\textcolor{red}{$\Delta$ \textit{Improvement}} & \textcolor{red}{+1.24} \\
\bottomrule
\end{tabular}
\caption{Ablation study of crop squarification for UI-Venus-7B on ScreenSpot Pro. }
\label{tab:ablation_squarification}
\end{minipage}

\end{table}

\paragraph{Using a square crop improves accuracy by preserving visual context.} UI elements vary widely in aspect ratio, and highly elongated crops can cause VLMs to misinterpret the spatial layout. By enforcing a square aspect ratio on the crop window, we observe a consistent improvement of +1.24 percentage points (60.56\% $\to$ 61.80\%, Table~\ref{tab:ablation_squarification}). This suggests that a compact, near-square crop better preserves the visual context needed for fine-grained grounding, leading to improved performance.

\subsection{Analysis}

\subsubsection{Analysis of Gating Threshold.}

\begin{figure}[!t]
    \centering
    \includegraphics[width=1.0\textwidth]{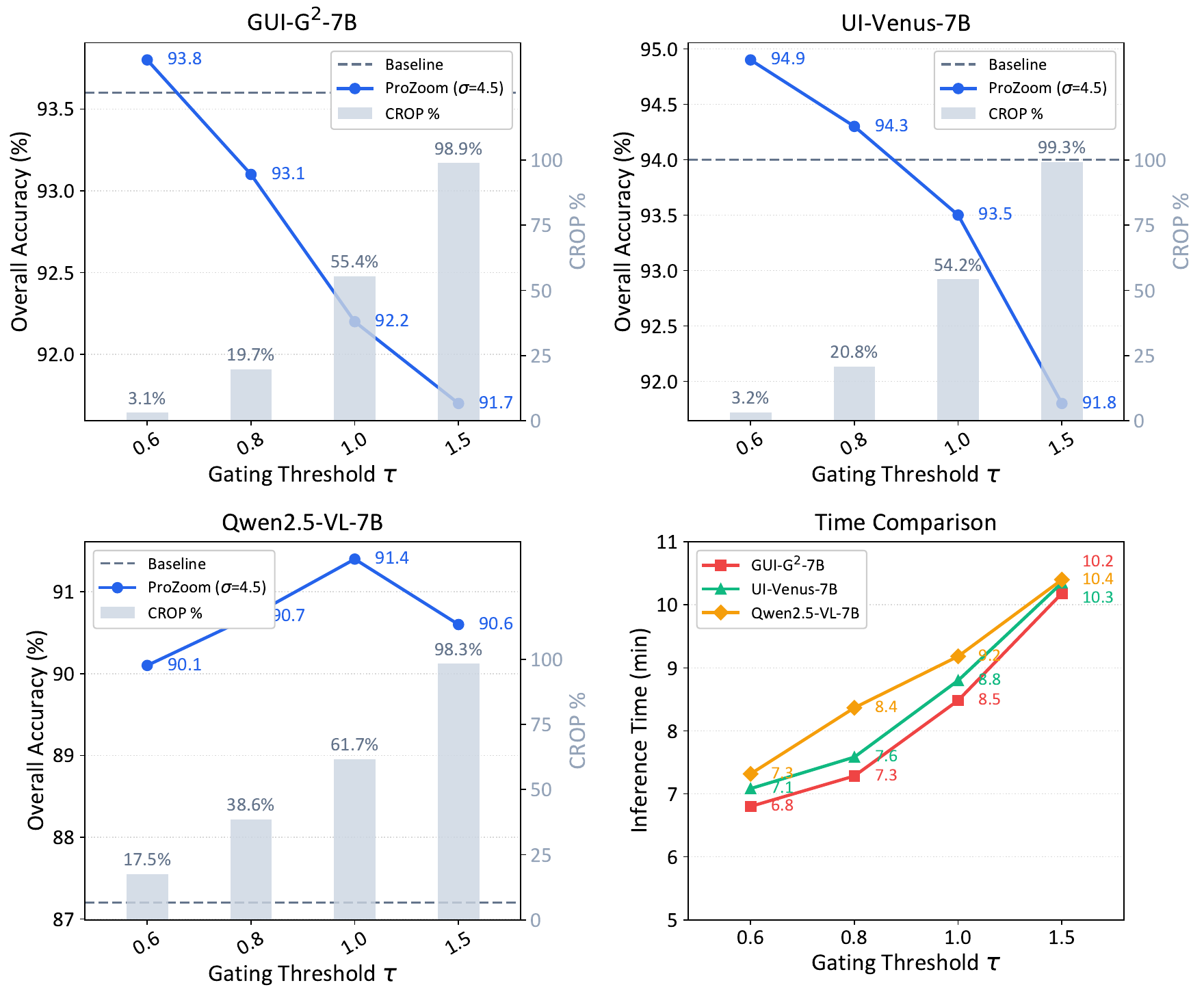}
    \caption{Ablation study of the gating threshold $\tau$ and Gaussian spread $\sigma$ on ScreenSpot-v2. Grey bars indicate the proportion of samples routed to the zoom-in cropping stage (CROP\%), while blue curves show overall accuracy.}
    \label{fig:threshold_ablation}
\end{figure}

To understand how the confidence-based gating threshold $\tau$ controls the trade-off between direct prediction and adaptive cropping, we ablate $\tau$ across three base models on ScreenSpot-v2, with $\sigma=4.5$ fixed. As shown in Figure~\ref{fig:threshold_ablation}, where $\sigma$ controls the size of the Gaussian-modeled crop window and CROP\% denotes the fraction of samples routed to the zoom-in stage, we draw three key observations:
\noindent\textbf{(1) Moderate thresholds yield the best accuracy.} When $\tau$ is too high, nearly all samples are cropped regardless of difficulty, hurting easy cases; when $\tau$ is too low, the method degenerates to the baseline. The optimal $\tau$ lies in between, selectively zooming in only when the model is uncertain.
\noindent\textbf{(2) Neither direct prediction nor full cropping alone is sufficient.} The baseline (CROP\%=0) leaves hard samples unresolved. Conversely, routing nearly all samples to the zoom-in stage (CROP\%$\approx$100\%) not only nearly doubles inference time (from $\sim$5:50 to $\sim$10:20) but also \textit{degrades} accuracy below the baseline (Figure~\ref{fig:threshold_ablation}), suggesting that indiscriminate cropping introduces noise rather than improving localization. Our gating mechanism bridges this gap, consistently outperforming both extremes across all three models while keeping computational overhead minimal.
\noindent\textbf{(3) Desktop and Web benefit more than Mobile.} Compared to Mobile, Desktop and Web interfaces contain denser layouts and smaller interactive elements, making them more sensitive to spatial ambiguity. \methodname's zoom-in stage provides finer local context that is especially effective in these environments.

\subsubsection{Analysis of Gating Signal Reliability.}

\begin{figure}[t]
    \centering
    \includegraphics[width=1.0\linewidth]{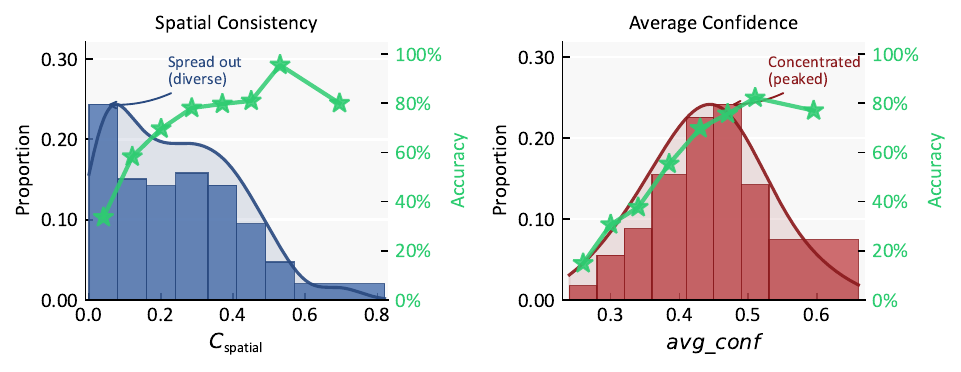}
    \caption{Histogram distributions of $C_{\mathrm{spatial}}$ and $avg\_conf$ on ScreenSpot-Pro ($N=1581$). The two signals exhibit complementary distributional characteristics, enabling a more discriminative gating mechanism when combined.}
    \label{fig:gating_hist}
\end{figure}

To verify that our two gating signals reliably reflect prediction confidence,
we bin all ScreenSpot-Pro samples ($N=1581$) by $C_{\mathrm{spatial}}$ and
$avg\_conf$ respectively and measure per-bin accuracy.
As shown in Figure~\ref{fig:gating_hist}, both signals show a general
positive correlation with accuracy, suggesting they serve as reasonable
proxies for localization reliability.
Furthermore, the two signals exhibit complementary distributional
characteristics: $C_{\mathrm{spatial}}$ is broadly spread while $avg\_conf$
is more concentrated, indicating that each captures a different aspect of
prediction uncertainty. Their combination thus yields a more discriminative
gating score, as corroborated by the ablation in Table~\ref{tab:gating_conpone_ablation}.

\subsubsection{Analysis of Sampling Number and Temperature.}
The effectiveness of \methodname hinges on the quality and diversity of the sampled candidate set, governed by sampling temperature $T$ and rollout count $N$. We ablate both on ScreenSpot-Pro (Figure~\ref{fig:TN_ablation}) and draw two conclusions: (1) \textbf{High temperature ($T{=}0.9$) is optimal.} Accuracy rises steadily from $54.46\%$ at $T{=}0.1$ to a peak of $\mathbf{61.80\%}$ at $T{=}0.9$, then marginally drops at $T{=}1.0$. This suggests that GUI grounding benefits from high candidate diversity: since consensus crop estimation relies on the spatial spread of predictions, diverse candidates better cover the true target region than conservative, near-identical ones. (2) \textbf{$N{=}8$ strikes the best accuracy--efficiency trade-off.} Accuracy peaks at $\mathbf{61.80\%}$ with $N{=}8$ and slightly declines for $N{=}12$ and $N{=}16$, as additional candidates beyond this point contribute redundant or noisy predictions that corrupt the crop estimation rather than refining it. We adopt $N{=}8$ as default.

\begin{figure}[!t]
    \centering
    \includegraphics[width=1.0\textwidth]{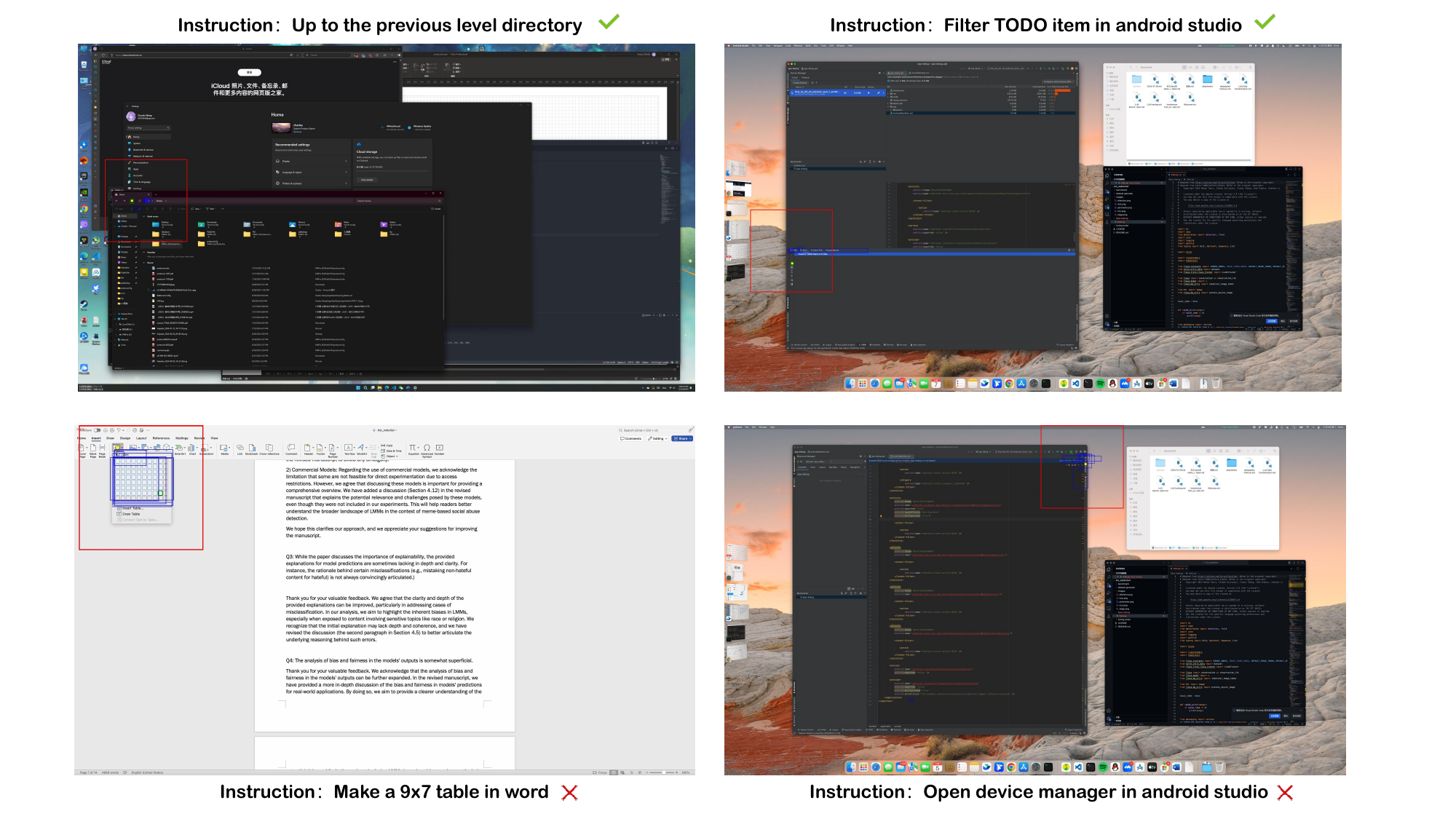}
    \caption{Here are four representative cases: the top two are successful examples, while the remaining two illustrate failure cases.
    Blue boxes denote the eight sampled bounding boxes, red indicates the cropped region, green represents the ground-truth box, and yellow marks the final prediction obtained after zooming on the cropped image.}
    \label{fig:case_study}
\end{figure}

\subsection{Case Studies}

To better understand \methodname's behavior beyond aggregate metrics, we visualize representative success and failure cases in Figure~\ref{fig:case_study}, where blue boxes denote the N stochastic candidate boxes from global multi-sampling, red highlights the zoom-in crop region, green is the ground-truth box, and yellow marks the final prediction.

In successful cases, \textbf{\methodname effectively identifies the correct target despite scattered initial predictions.} Even when the initial candidates are dispersed and none precisely overlap the target, \methodname leverages the spatial distribution of these proposals to identify a reliable crop region. The model then refines the prediction with a single zoom-in pass, locking onto the correct UI element. This demonstrates the method's robustness in handling difficult cases, where initial uncertainty is high, but the model can still correctly localize the target. 

In failure cases, \textbf{strong visual distractors and ambiguous cues lead to incorrect predictions.} These scenarios often involve multiple similar-looking icons in dense layouts, with the true target being extremely small and difficult to distinguish. In such cases, \methodname struggles to resolve the ambiguity and accurately identify the correct target, illustrating the challenges posed by highly cluttered interfaces.

\section{Conclusion}
We present \methodname, a adaptive zoom-in framework for GUI grounding that treats both the trigger and scale of zoom-in as a prediction uncertainty quantification problem. By fusing spatial consensus with token-level confidence, our reliability gate selectively routes uncertain instances to an adaptive cropping stage, where the crop window is derived from a principled variance decomposition. Extensive experiments on ScreenSpot-Pro, UI-Vision, and ScreenSpot-v2 demonstrate consistent improvements across four model architectures, with gains of up to +13.4\%, +10.3\%, and +4.2\%, respectively. UI-Zoomer establishes that zoom only when uncertain, and zoom by how much the predictions disagree is a simple yet effective principle for test-time scaling in GUI grounding.

\bibliography{main,custom}
\bibliographystyle{main}

\clearpage

\tableofcontents

\clearpage

\appendix
\section{Appendix}
This appendix provides additional implementation and evaluation details to complement the main paper. 

\subsection{Prompt Template}
\begin{figure}[htbp]
    \centering
    \includegraphics[width=0.8\linewidth]{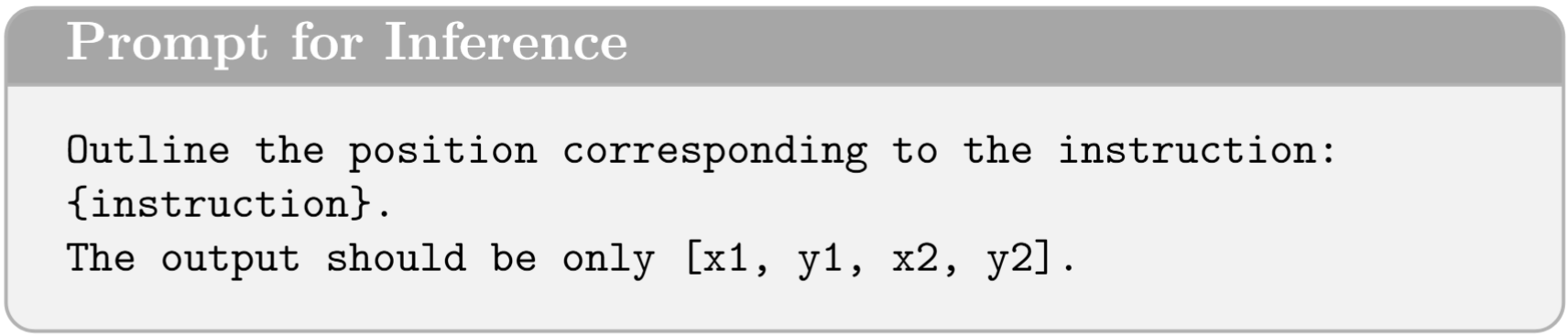}
    \caption{Complete prompt template used in our experiments.}
    \label{fig:prompt}
\end{figure}

Our experiments adopt a unified prompt template for inference, as illustrated in Figure \ref{fig:prompt}. We use this prompt consistently across all experimental settings to ensure a fair comparison among different models and evaluation scenarios.

\subsection{Comprehensive Comparison on ScreenSpot-v2 and UI-Vision}  
Table \ref{tab:ssv2_results} and Table \ref{tab:ui_vision_results} present the complete experimental results of our proposed \methodname on the \textbf{ScreenSpot-v2} and \textbf{UI-Vision} benchmarks, respectively, together with comparisons against a broad range of existing baselines. These results provide a comprehensive evaluation of our method across different models, environments, and task settings. Overall, the comparisons show that \methodname consistently improves localization performance over the corresponding base models and achieves competitive or superior results relative to existing methods, demonstrating its effectiveness and generalizability across diverse UI grounding benchmarks.

\begin{table*}[!t]
\centering
\resizebox{\textwidth}{!}{%
\begin{tabular}{l|ccc|ccc|ccc|ccc}
\toprule
\multirow{2}{*}{\textbf{Methods}} &
\multicolumn{3}{c|}{\textbf{Mobile}} & 
\multicolumn{3}{c|}{\textbf{Desktop}} & 
\multicolumn{3}{c|}{\textbf{Web}} & 
\multicolumn{3}{c}{\textbf{Overall}} \\
 & text & icon & avg & text & icon & avg & text & icon & avg & text & icon & avg \\
\midrule

\rowcolor{gray!15}
\multicolumn{13}{l}{\textit{Proprietary Methods}} \\
GPT-4o~\cite{gpt4o} & - & - & 22.5 & - & - & 22.2 & - & - & 12.4 & - & - & 20.1 \\
Claude-3.7-Sonnet~\cite{claude37sonnet} & - & - & - & - & - & - & - & - & - & - & - & 87.6 \\
Seed-1.5-VL~\cite{seed15vl} & - & - & - & - & - & - & - & - & - & - & - & 95.2 \\

\midrule
\rowcolor{gray!15}
\multicolumn{13}{l}{\textit{General Open-source Models}} \\
OS-Atlas-4B~\cite{osatlas} & 87.2 & 59.7 & 74.9 & 72.7 & 46.4 & 56.9 & 85.9 & 63.1 & 70.0 & - & - & 71.9 \\
OS-Atlas-7B & 95.2 & 75.8 & 78.3 & 90.7 & 63.6 & 85.5 & 90.6 & 77.3 & 83.8 & - & - & 84.1 \\
Qwen2.5-VL-3B~\cite{qwen25vl} & 93.4 & 73.5 & - & 88.1 & 58.6 & - & 88.0 & 71.4 & - & - & - & 80.9 \\ 
Qwen2.5-VL-32B & 98.3 & 86.7 & - & 94.3 & 83.6 & - & 93.6 & 89.7 & - & - & - & 91.9 \\ 
UGround-7B~\cite{uground} & - & - & 74.3 & - & - & 74.9 & - & - & 78.6 & - & - & 76.3 \\
UI-TARS-7B & 96.9 & 89.1 & - & 95.4 & 85.0 & - & 93.6 & 85.2 & - & - & - & 91.6 \\
UI-TARS-72B &  94.8 & 86.3 & - & 91.2 & 87.9 & - & 91.5 & 87.7 & - & - & - & 90.3 \\
Jedi-3B~\cite{jedi} & 96.6 & 81.5 & - & 96.9 & 78.6 & - & 88.5 & 83.7 & - & - & - & 88.6 \\
Jedi-7B & 96.9 & 87.2 & - & 95.9 & 87.9 & - & 94.4 & 84.2 & - & - & - & 91.7 \\

\midrule
\rowcolor{gray!15}
\multicolumn{13}{l}{\textit{Reinforcement Learning Methods}} \\
UI-TARS-1.5-7B~\cite{uitars15} & 96.5 & 86.4 & - & 95.0 & 87.3 & - & 88.2 & 86.5 & - & - & - & 90.2 \\
GTA1-3B~\cite{gta1} & 99.0 & 88.6 & - & 94.9 & 89.3 & - & 92.3 & 86.7 & - & - & - & 92.4 \\ 
GTA1-7B & 99.7 & 90.5 & - & 99.0 & 94.3 & - & 95.7 & 90.1 & - & - & - & 95.2 \\
UI-R1-E-3B~\cite{uir1} & 98.2 & 83.9 & - & 93.2 & 83.7 & - & 94.8 & 75.0 & - & - & - & 89.5 \\ 
UI-S1-7B~\cite{u1s1} & - & - & - & - & - & - & - & - & - & - & - & 90.1 \\
SE-GUI-7B~\cite{segui} & - & - & 95.2 & - & - & 87.1 & - & - & 87.0 & - & - & 90.3 \\

\midrule
\rowcolor{gray!15}
\multicolumn{13}{l}{\textit{Test Scaling Methods}} \\
DiMo-GUI~\cite{dimo} & 94.8 & 85.3 & - & 94.3 & 82.1 & - & 93.2 & 80.3 & - & - & - & 89.2 \\
GUI-RC~\cite{guirc} & 99.9 & 85.9 & - & 91.1 & 73.0 & - & 91.8 & 81.4 & - & - & - & 88.5 \\

\midrule
\rowcolor{gray!15}
\multicolumn{13}{l}{\textit{Our method}} \\

UI-Venus-7B & 
99.0 & 90.1 & 95.2 & 
97.9 & 89.3 & 94.3 & 
94.9 & 89.7 & 92.5 & 
97.4 & 89.7 & 94.0 \\
\textbf{+ \methodname} & 
\textbf{98.6} & \textbf{90.5} & \textbf{95.2} & 
\textbf{99.0} & \textbf{92.9} & \textbf{96.4} & 
\textbf{95.7} & \textbf{90.6} & \textbf{93.4} & 
\textbf{97.8} & \textbf{91.2} & \textbf{94.9} \\
\textcolor{red}{$\Delta$ \textit{Improvement}} & 
\textcolor{red}{-0.4} & \textcolor{red}{+0.4} & \textcolor{red}{+0.0} & 
\textcolor{red}{+1.1} & \textcolor{red}{+3.6} & \textcolor{red}{+2.1} & 
\textcolor{red}{+0.8} & \textcolor{red}{+0.9} & \textcolor{red}{+0.9} & 
\textcolor{red}{+0.4} & \textcolor{red}{+1.5} & \textcolor{red}{+0.9} \\
\midrule
\midrule

Qwen2.5-VL-7B & 
97.2 & 86.7 & 92.8 & 
87.6 & 67.1 & 79.0 & 
90.2 & 83.3 & 87.0 & 
92.3 & 80.5 & 87.2 \\
\textbf{+ \methodname} & 
\textbf{97.9} & \textbf{87.2} & \textbf{93.4} & 
\textbf{97.9} & \textbf{84.3} & \textbf{92.2} & 
\textbf{91.0} & \textbf{85.7} & \textbf{88.6} & 
\textbf{95.7} & \textbf{85.9} & \textbf{91.4} \\
\textcolor{red}{$\Delta$ \textit{Improvement}} & 
\textcolor{red}{+0.7} & \textcolor{red}{+0.5} & \textcolor{red}{+0.6} & 
\textcolor{red}{+10.3} & \textcolor{red}{+17.2} & \textcolor{red}{+13.2} & 
\textcolor{red}{+0.8} & \textcolor{red}{+2.4} & \textcolor{red}{+1.6} & 
\textcolor{red}{+3.4} & \textcolor{red}{+5.4} & \textcolor{red}{+4.2} \\
\midrule
\midrule

GUI-G$^2$-7B & 
99.3 & 91.5 & 96.0 & 
97.9 & 85.7 & 92.8 & 
93.6 & 89.2 & 91.5 & 
97.1 & 89.2 & 93.6 \\
\textbf{+ \methodname} & 
\textbf{98.6} & \textbf{91.5} & \textbf{95.6} & 
\textbf{97.9} & \textbf{91.4} & \textbf{95.2} & 
\textbf{92.7} & \textbf{88.2} & \textbf{90.6} & 
\textbf{96.5} & \textbf{90.3} & \textbf{93.8} \\
\textcolor{red}{$\Delta$ \textit{Improvement}} & 
\textcolor{red}{-0.7} & \textcolor{red}{+0.0} & \textcolor{red}{-0.4} & 
\textcolor{red}{+0.0} & \textcolor{red}{+5.7} & \textcolor{red}{+2.4} & 
\textcolor{red}{-0.9} & \textcolor{red}{-1.0} & \textcolor{red}{-0.9} & 
\textcolor{red}{-0.6} & \textcolor{red}{+1.1} & \textcolor{red}{+0.2} \\

\bottomrule
\end{tabular}%
}
\caption{Performance comparison on ScreenSpot-v2. We evaluate our \methodname strategy across three models: Qwen2.5-VL-7B, GUI-G$^2$-7B, and UI-Venus-7B. }
\label{tab:ssv2_results}
\end{table*}

\begin{table}[!t]
\centering
\resizebox{\textwidth}{!}{%
\begin{tabular}{l|ccc|ccc|ccc|ccc}
\toprule
\multirow{2}{*}{\textbf{Methods}} &
\multicolumn{3}{c|}{\textbf{Basic}} & 
\multicolumn{3}{c|}{\textbf{Functional}} & 
\multicolumn{3}{c|}{\textbf{Spatial}} & 
\multicolumn{3}{c}{\textbf{Overall}} \\
 & text & icon & avg & text & icon & avg & text & icon & avg & text & icon & avg \\
\midrule

\rowcolor{gray!15}
\multicolumn{13}{l}{\textit{Proprietary Methods}} \\
GPT-4o~\cite{gpt4o} & - & - & 1.6 & - & - & 1.5 & - & - & 1.0 & - & - & 1.4 \\
Claude-3.7-Sonnet~\cite{claude37sonnet} & - & - &  9.5 & - & - &  7.7 & - & - & 7.6 & - & - & 8.3 \\
Seed-1.5-VL~\cite{seed15vl} & - & - & - & - & - & - & - & - & - & - & - & - \\

\midrule
\rowcolor{gray!15}
\multicolumn{13}{l}{\textit{General Open-source Models}} \\
OS-Atlas-7B~\cite{osatlas} & - & - & 12.2 & - & - & 11.2 & - & - & 3.7 & - & - & 9.0 \\
UGround-7B~\cite{uground} & - & - &  11.5 & - & - & 12.2 & - & - & 2.8 & - & - & 8.8 \\
UGround-72B & - & - &  27.9 & - & - &  26.7 & - & - & 14.9 & - & - &  23.2 \\
UI-TARS-7B & - & - & 20.1 & - & - &  24.3 & - & - & 8.4 & - & - & 17.6 \\
UI-TARS-72B & - & - & 31.4 & - & - & 30.5 & - & - & 14.7 & - & - & 25.5 \\
Jedi-3B~\cite{jedi} & - & - & 22.3 & - & - & 25.2 & - & - & 9.4 & - & - & 18.7 \\
Jedi-7B & - & - & 32.3 & - & - & 30.5 & - & - & 12.8 & - & - & 24.8 \\

\midrule
\rowcolor{gray!15}
\multicolumn{13}{l}{\textit{Reinforcement Learning Methods}} \\
UI-TARS-1.5~\cite{uitars15} & - & - & 28.8 & - & - & 27.5 & - & - & 10.7 & - & - & 22.3 \\
InfiGUI-G1-3B~\cite{infig1}& - & - & 31.2 & - & - & 28.0 & - & - & 8.2 & - & - & 22.0 \\
InfiGUI-G1-7B& - & - & 36.2 & - & - & 31.9 & - & - & 11.5 & - & - & 26.1 \\

\midrule
\rowcolor{gray!15}
\multicolumn{13}{l}{\textit{Our method}} \\

Qwen2.5-VL-7B & 
57.3 & 10.7 & 20.6 & 
40.2 & 10.0 & 16.5 & 
18.9 & 4.8 & 6.4 & 
38.2 & 7.9 & 13.3 \\
\textbf{+ \methodname} & 
\textbf{75.7} & \textbf{22.2} & \textbf{33.6} & 
\textbf{59.8} & \textbf{19.7} & \textbf{28.4} & 
\textbf{38.7} & \textbf{9.3} & \textbf{12.7} & 
\textbf{57.0} & \textbf{16.3} & \textbf{23.6} \\
\textcolor{red}{$\Delta$ \textit{Improvement}} & 
\textcolor{red}{+18.4} & \textcolor{red}{+11.5} & \textcolor{red}{+13.0} & 
\textcolor{red}{+19.6} & \textcolor{red}{+9.7} & \textcolor{red}{+11.9} & 
\textcolor{red}{+19.8} & \textcolor{red}{+4.5} & \textcolor{red}{+6.3} & 
\textcolor{red}{+18.8} & \textcolor{red}{+8.4} & \textcolor{red}{+10.3} \\

\midrule
\midrule

GUI-G$^2$-7B & 
66.7 & 24.0 & 33.0 & 
65.8 & 21.9 & 31.4 & 
36.5 & 7.7 & 11.0 & 
59.5 & 17.1 & 24.7 \\
\textbf{+ \methodname} & 
\textbf{78.4} & \textbf{32.6} & \textbf{42.3} & 
\textbf{69.5} & \textbf{30.9} & \textbf{39.2} & 
\textbf{53.6} & \textbf{14.0} & \textbf{18.6} & 
\textbf{69.3} & \textbf{25.0} & \textbf{32.9} \\
\textcolor{red}{$\Delta$ \textit{Improvement}} & 
\textcolor{red}{+11.7} & \textcolor{red}{+8.6} & \textcolor{red}{+9.3} & 
\textcolor{red}{+3.7} & \textcolor{red}{+9.0} & \textcolor{red}{+7.8} & 
\textcolor{red}{+17.1} & \textcolor{red}{+6.3} & \textcolor{red}{+7.6} & 
\textcolor{red}{+9.8} & \textcolor{red}{+7.9} & \textcolor{red}{+8.2} \\

\midrule
\midrule

UI-Venus-7B & 
68.5 & 23.8 & 33.3 & 
65.5 & 19.9 & 29.7 & 
38.7 & 7.8 & 11.4 & 
60.6 & 16.5 & 24.4 \\
\textbf{+ \methodname} & 
\textbf{80.5} & \textbf{32.3} & \textbf{42.5} & 
\textbf{74.7} & \textbf{30.6} & \textbf{40.1} & 
\textbf{55.9} & \textbf{15.1} & \textbf{19.7} & 
\textbf{72.7} & \textbf{25.2} & \textbf{33.7} \\
\textcolor{red}{$\Delta$ \textit{Improvement}} & 
\textcolor{red}{+12.0} & \textcolor{red}{+8.5} & \textcolor{red}{+9.2} & 
\textcolor{red}{+9.2} & \textcolor{red}{+10.7} & \textcolor{red}{+10.4} & 
\textcolor{red}{+17.2} & \textcolor{red}{+7.3} & \textcolor{red}{+8.3} & 
\textcolor{red}{+12.1} & \textcolor{red}{+8.7} & \textcolor{red}{+9.3} \\

\midrule
\midrule

UI-Venus-72B & 
72.1 & 30.6 & 39.4 & 
68.7 & 28.1 & 36.9 & 
43.5 & 12.8 & 16.3 & 
64.3 & 23.6 & 30.9 \\
\textbf{+ \methodname} & 
\textbf{83.2} & \textbf{39.0} & \textbf{48.4} & 
\textbf{77.8} & \textbf{34.9} & \textbf{44.2} & 
\textbf{67.6} & \textbf{24.8} & \textbf{29.7} & 
\textbf{77.6} & \textbf{32.3} & \textbf{40.4} \\
\textcolor{red}{$\Delta$ \textit{Improvement}} & 
\textcolor{red}{+11.1} & \textcolor{red}{+8.4} & \textcolor{red}{+9.0} & 
\textcolor{red}{+9.1} & \textcolor{red}{+6.8} & \textcolor{red}{+7.3} & 
\textcolor{red}{+24.1} & \textcolor{red}{+12.0} & \textcolor{red}{+13.4} & 
\textcolor{red}{+13.3} & \textcolor{red}{+8.7} & \textcolor{red}{+9.5} \\

\bottomrule
\end{tabular}%
}
\caption{Performance comparison on UI-Vision. We evaluate our \methodname strategy across four models: Qwen2.5-VL-7B, GUI-G$^2$-7B, UI-Venus-7B, and UI-Venus-72B .}
\label{tab:ui_vision_results}
\end{table}

\subsection{More Ablations}
The results demonstrated in Table \ref{tab:gating_rationality} validate the rationality of our gating mechanism: samples routed to the \textit{Gating Pass} branch consistently exhibit significantly higher accuracy than those sent to the \textit{Crop} branch, demonstrating that the gating score $S$ reliably reflects prediction confidence.

\begin{table}[htbp]
\centering
\small
\begin{tabular}{lc|cc|cc}
\toprule
\multirow{2}{*}{\textbf{Threshold} ($\tau$)} & \multirow{2}{*}{\textbf{Overall Acc}} & \multicolumn{2}{c|}{\textbf{Gating Pass Branch}} & \multicolumn{2}{c}{\textbf{Crop Branch}} \\
\cmidrule{3-6}
& & \textbf{Trigger Rate} & \textbf{Accuracy} & \textbf{Trigger Rate} & \textbf{Accuracy} \\
\midrule
0.6  & 57.56\% & 53.89\% & 70.66\% & 46.11\% & 42.24\% \\
0.8  & 60.97\% & 22.77\% & 79.44\% & 77.23\% & 33.42\% \\
0.9  & 61.48\% & 9.93\%  & 82.80\% & 90.07\% & 59.13\% \\
0.95 & 61.54\% & 5.57\%  & 81.82\% & 94.43\% & 60.34\% \\
1.0  & \textbf{61.80\%} & 2.91\%  & 93.48\% & 97.09\% & 60.85\% \\
1.05 & \textbf{61.80\%} & 1.64\%  & \textbf{96.15\%} & 98.36\% & 61.22\% \\
\bottomrule
\end{tabular}
\caption{Ablation of the Gating Threshold ($\tau$). These are the results of UI-Venus-7B on ScreenSpot Pro, with the hyperparameter $\sigma$ is set to 2.5.}
\label{tab:gating_rationality}
\end{table}

\clearpage

\end{document}